# DESIGN OF LOW NOISE MICROWAVE AMPLIFIERS USING PARTICLE SWARM OPTIMIZATION

Sadık Ülker

Department of Electrical and Electronics Engineering, Girne American University, Girne, Mersin-10, Turkey
sulker@gau.edu.tr

### ABSTRACT

*This short paper presents a work on the design of low noise microwave amplifiers using particle swarm optimization (PSO) technique. Particle Swarm Optimization is used as a method that is applied to a single stage amplifier circuit to meet two criteria: desired gain and desired low noise. The aim is to get the best optimized design using the predefined constraints for gain and low noise values. The code is written to apply the algorithm to meet the desired goals and the obtained results are verified using different simulators. The results obtained show that PSO can be applied very efficiently for this kind of design problems with multiple constraints.*

## 1. INTRODUCTION

Particle Swarm Optimization, PSO, is a stochastic optimization technique which was developed by Kennedy and Eberhart in 1995, inspired by the behavior of bird flocks, or bee swarms [1]. It is a population based optimization tool and can be implemented and applied very easily to solve various function optimization problems in engineering and science. Since the discovery of particle swarm optimization many researchers worked on the improvement of the method by analyzing the algorithm performance using modifications in the parameters changing them [2]-[3]. Over the past few years some researchers worked on the application of PSO to different optimization problems and it has also recently been started to gain an increasing interest and use in electronic circuit design, antenna design and microwave applications as an optimizing and design tool [4]-[10]. In this work, to the author's knowledge, the method is applied for the design of low noise microwave transistor amplifiers for the first time.

## 2. DESIGN

### 2.1. General Design Approach

The main optimization process for particle swarm optimization algorithm is well known and can be described with a pseudo code below:

**-Initialize particles**
**-Calculate the fitness value for each particle**
**-Do while the minimum error criteria is not reached**
    **-Check the fitness value for each particle, replace *pbest* if fitness value is better than the current *pbest* of the particle,**
    **-Among the particles, choose the particle with the best *pbest* and label this as the *gbest*,**
    -Calculate and update each particles new velocity and position with the equations **(1) and (2)**.

The first step in the optimization process is to initialize a group of particles in a random manner. For each particle, at every iteration, fitness values, which show how close the particles are to the desired solution, are calculated. In every iteration, for each particle there is a local best solution (pbest) that is achieved so far in iterations, and among all the particles there is a global best solution (gbest). According to the equations

$$v[] = w*v[] + c*rand()*(pbest[] - present[]) + c*rand()*(gbest[] - present[]) \quad (1)$$
$$present[] = present[] + v[] \quad (2)$$

the particle updates its velocity, v[], and position, present[], and iterations continue until all particles reach to an optimized solution. In this work, inertia weight *w*, is set to 0.4 and learning parameter *c*, is set to 2.00. In this work, these values gave the best convergence to an answer compared with other values. *rand()* is a random number and it gets generated by computer for a range of values between 0 and 1 randomly at every iteration for each particle.

## 2.2. Low Noise Microwave Amplifier Design

Low noise microwave amplifier is a very important component in receiver applications especially when it is required to have preamplifier with as low a noise figure as possible and details of a low noise amplifier design can be found in many texts [11]-[14]. A detailed explanation for CMOS low noise amplifier design with different optimisation techniques was presented by Nguyen et. al. [15]. Generally it is not possible to obtain both maximum gain and minimum noise figure therefore a trade off exists between noise figure and gain. Therefore the task is essentially solving a multiobjective optimization problem. Similar work, designing a CMOS operational transconductance amplifier, was done by converting the constrained optimisation problem into an unconstrained one by Bennour et. al. [16]. Recently memetic algorithm was applied to design a low noise microwave amplifier by some researchers [17]. In this work, microwave low noise transistor amplifier design is done using particle swarm optimization technique.

The general microwave single stage transistor amplifier circuit is shown in Figure 1 [18]. Transducer power gain ($G_T$) which is defined to be the ratio of power delivered to the load, to the power available from the source is given by the equation:

$$G_T = \frac{|S_{21}|^2 (1 - |\Gamma_S|^2)(1 - |\Gamma_L|^2)}{|(1 - S_{11}\Gamma_S)(1 - S_{22}\Gamma_L) - (S_{12}S_{21}\Gamma_S\Gamma_L)|^2} \quad (3)$$

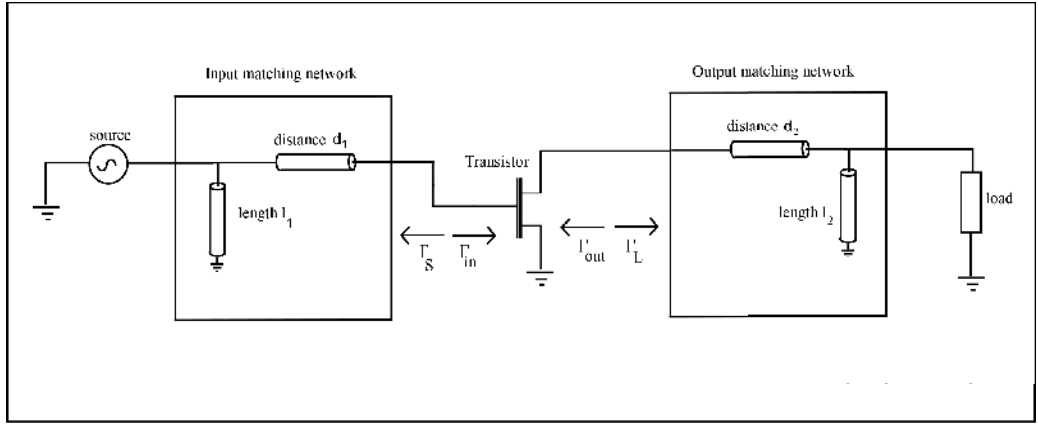

Figure 1. Single Stage Microwave Amplifier Circuit

In the optimization procedure the value for desired gain is calculated using equation (3) and this equation is one of the equations that used in determining whether the obtained values are close to their optimized value. In equation (3) $\Gamma_S$ is the reflection coefficient looking toward the source, $\Gamma_L$ is the reflection coefficient looking toward the load and $S_{11}$, $S_{12}$, $S_{21}$, $S_{22}$ are the transistor s-parameters. Since the s-parameters of the device is fixed for a specific frequency and operating condition, the design becomes finding proper $\Gamma_S$ and $\Gamma_L$ values and hence obtaining the proper load and source impedances so that the desired gain can be obtained. For the single stage microwave amplifier circuit shown in Figure 1, the relationships for $\Gamma_S$ and $\Gamma_L$ are related to the transmission line lengths ($d_1$, $d_2$, $l_1$ and $l_2$) and can be obtained via these equations [15]:

$$\Gamma_S = \frac{Zsource - 1}{Zsource + 1} \quad (4)$$

$$\Gamma_L = \frac{Zload - 1}{Zload + 1} \quad (5)$$

where

$$Zsource = \frac{\text{Re}(Zprs) + j(Zo \tan(2\pi d_1) + \text{Im}(Zprs))}{Zo - \text{Im}(Zprs) * \tan(2\pi d_1) + j \text{Re}(Zprs) * \tan(2\pi d_1)} \quad (6)$$

$$Zprs = \frac{jZo \tan(2\pi l_1)}{1 + j \tan(2\pi l_1)} \quad (7)$$

and similarly

$$Zload = \frac{\text{Re}(Zprl) + j(Zo \tan(2\pi d_2) + \text{Im}(Zprl))}{Zo - \text{Im}(Zprl) * \tan(2\pi d_2) + j \text{Re}(Zprl) * \tan(2\pi d_2)} \quad (8)$$

$$Zprl = \frac{jZo \tan(2\pi l_2)}{1 + j \tan(2\pi l_2)} \quad (9)$$

In the design, the desired gain value is needed to be optimized at a single frequency. In order to do this, a search for the length parameters directly which will satisfy the desired gain at the desired predetermined frequency is done. For stability, the overall systems input and output reflection coefficient magnitudes are always checked to be below 0.99, which can be considered as an additional constraint. Besides this, another criterion is considered which is the noise figure value being below the maximum predetermined limit. With these constraints, the problem can be considered as an optimization problem with two inequality constraints as explained by Noceidal and Wrigth [19]. Noise figure relations are well known:

$$F = F_{\min} + \frac{R_N}{G_S} |Y_S - Y_{opt}|^2 \quad (10)$$

$$Y_S = G_S + jB_S \quad (11)$$

$$Y_S = \frac{1}{Z_o} \frac{1-\Gamma_S}{1+\Gamma_S} \quad (12)$$

$$Y_{opt} = \frac{1}{Z_o} \frac{1-\Gamma_{opt}}{1+\Gamma_{opt}} \quad (13)$$

where F is the noise figure, $F_{\min}$ is the minimum noise figure that can be attained with the transistor used, $\Gamma_{opt}$ is the reflection coefficient necessary for attaining the minimum noise figure predicted, and $R_N$ is the equivalent noise resistance of the transistor. $Y_{opt}$ is optimum source admittance that results in minimum noise figure, $Y_S$ is source admittance presented to the source, which is formed by $G_S$, real part of source admittance, and $B_S$, imaginary part of the source admittance. The parameters $F_{\min}$, $\Gamma_{opt}$ (and hence $Y_{opt}$), and $R_N$ are fixed since they are transistor characteristic and hence they are fixed for a specific design frequency under normal operating conditions. However F and $\Gamma_S$ will change with a change in lengths (i.e. particles). In this work, the gain, as given in (3), is used as a fitness check and the noise figure, F, as given in (10), is used as a fitness criteria like $\Gamma_{in}$ and $\Gamma_{out}$ for stability being in restricted values. Gain is tried to be optimized to a high desired value, with the criteria that the noise figure value being below the given limit at the design frequency.

## 3. WORK DONE

In the test circuits, Fujitsu Cooperation's high-electron mobility transistor (HEMT), FHX35X, was used in the test for designing amplifiers. A set of 15 'particles' which hold four values; the values of input matching networks distance and length and output matching networks distance and length, (i.e. lengths $d_1$, $d_2$, $l_1$ and $l_2$) were created randomly. The creation was done by means of a random number generator between the lengths 0 and 0.1 $\lambda_c$ ($\lambda_c$ being the wavelength for the design center frequency).

For the design, gain requirement was set to 20 dB and noise figure criterion was set to being below 1.0 dB. 3000 iterations were computed where all the particles reach to a solution. A sample graph showing the iteration convergence is shown in Figure 2. As we can see from the figure after 100 iterations, 3 of the 15 particles reach to an optimized solution and after about 500 iterations more than 12 of the 15 particles are within the limits for an optimized solution.

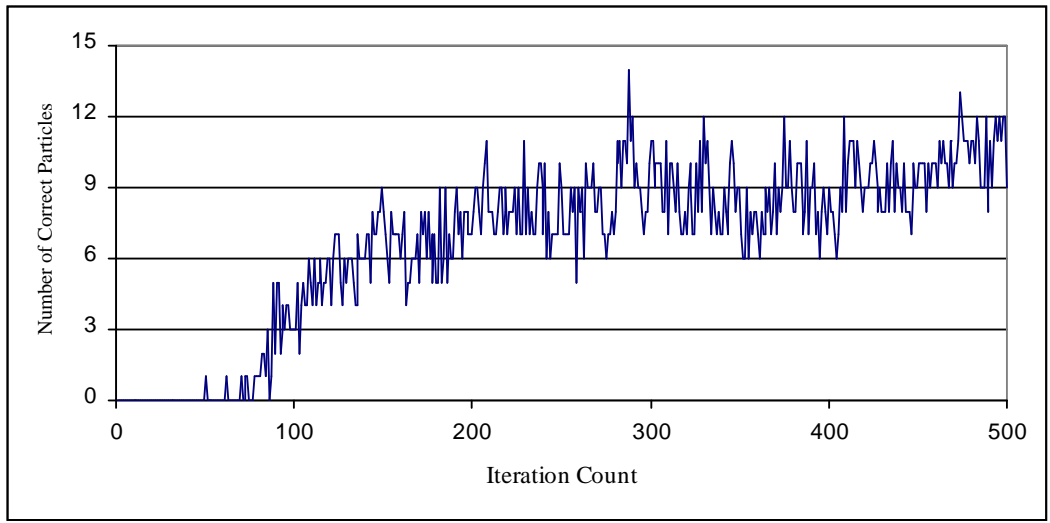

Figure 2. Convergence of particles to a desired solution vs. number of iterations

These values were used in PUFF [20] microwave circuit simulator for gain check and also (10) is used for checking whether the minimum noise criterion was met. Table 1 shows some of the results that were obtained with different program runs. We can see from the table the obtained lengths and corresponding $\Gamma_S$ values. We can draw the noise figure circle and locate the obtained source reflection coefficients, $\Gamma_S$ to see if the minimum noise criterion is met. This is shown in Figure 3. The $\Gamma_S$ values are marked as crosses and we can easily see that they are all within the 1dB noise figure circle. The gain response for the third trial is shown in Figure 4. Gain of 20.01 dB is achieved as designed, while keeping the overall system reflections below 0 dB: $|S_{11}|$= -2.50 dB, $|S_{22}|$= -2.73 dB. This also ensures that the amplifier is stable at the design frequency. Although the design seems to be working well at the design frequency, we can see that the system reflection coefficients at other frequencies were not very favourable in the sense that they were approaching to 0 dB at some frequencies. This can be eliminated via adding extra constraints at other frequencies to the program.

Table 1. Lengths and calculated Noise Figure

|  | $d_1$ | $l_1$ | $d_2$ | $l_2$ | $|\Gamma s|$ | $\Gamma s$ angle | Noise Figure (dB) |
|---|---|---|---|---|---|---|---|
| **Trial 1** | 41.1599° | 23.8961° | 34.3209° | 50.9232° | 0.748292 | 56.06473° | 0.847376 |
| **Trial 2** | 43.9317° | 20.6296° | 30.6803° | 61.7606° | 0.798747 | 55.08966° | 0.901598 |
| **Trial 3** | 45.1513° | 21.6466° | 16.1073° | 61.3262° | 0.783121 | 51.18767° | 0.714466 |
| **Trial 4** | 46.9831° | 20.7794° | 29.0269° | 56.6803° | 0.796434 | 48.84327° | 0.656502 |
| **Trial 5** | 41.771° | 23.9152° | 41.7792° | 43.972° | 0.74804 | 54.82775° | 0.800902 |

.

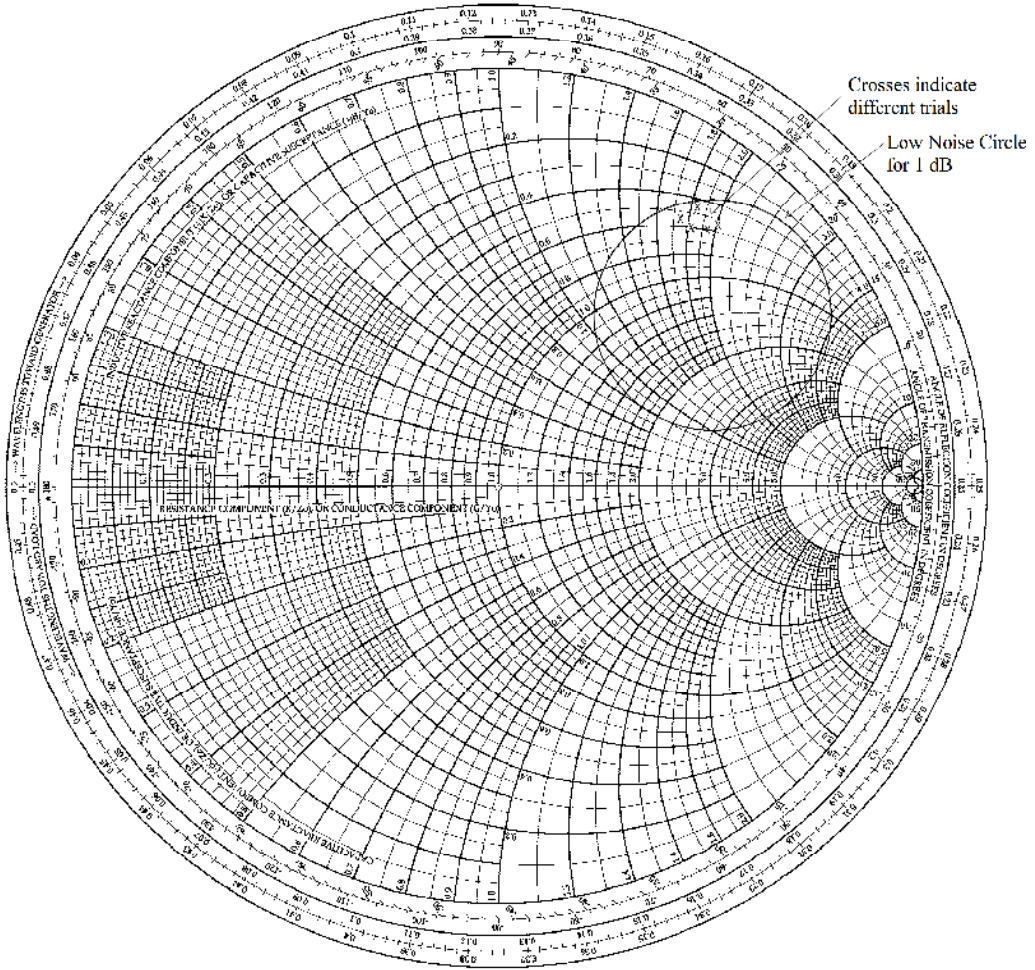

Figure 3. Design being shown on Smith Chart

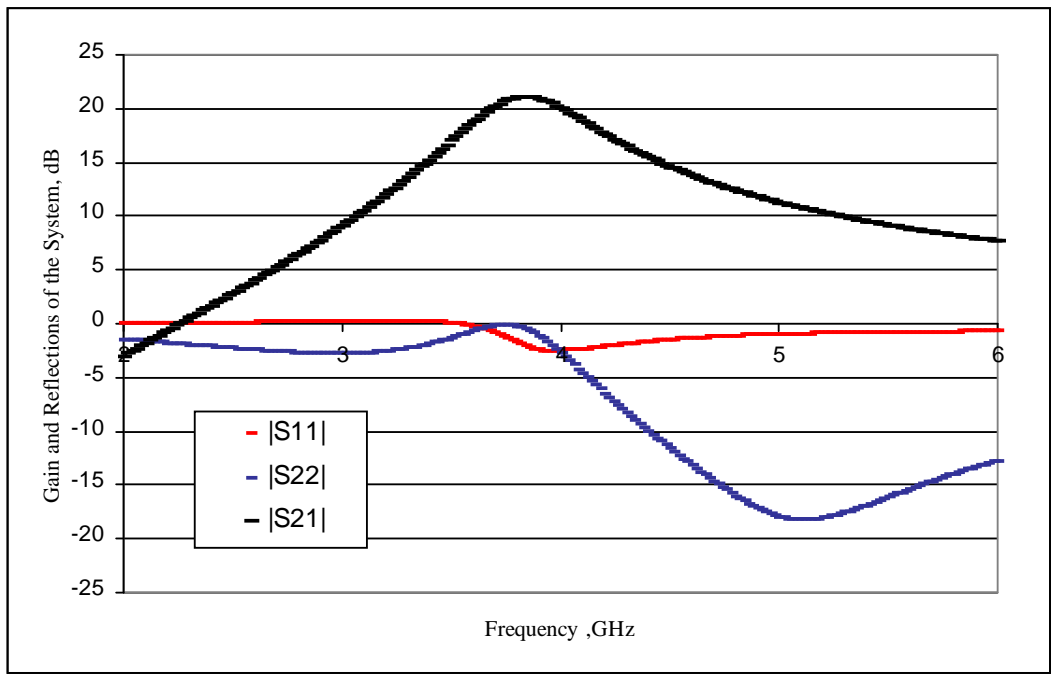

Figure 4. Amplifier Design completed showing the Gain and Reflection Responses

## 4. CONCLUSIONS

The aim of this work was to show that a design of low noise microwave transistor amplifier circuit, which can be thought of a multiobjective optimization problem, can be performed with the aid of particle swarm optimization. Design circuit architecture is based on a simple transmission line matching circuit architecture. The particles hold the values of these transmission line lengths and the search was done to find the values of these transmission line lengths to meet the desired gain and noise figure constraints. The results were obtained very fast, in less than 3000 iterations, and these design values obtained were used in a microwave simulator to verify their correctness. The response of the amplifier indicated that both the predetermined gain and low noise criteria are met. We can conclude that this successful application of the particle swarm optimization in design can lead the microwave circuit simulator writers to use particle swarm optimization as an algorithm in their simulator tools to optimize the possible design quantities.

## Author

**Sadik Ulker** received the B.Sc., M.E. and Ph.D. degree in electrical engineering from University of Virginia, Charlottesville, VA, USA, in 1996, 1999, and 2002 respectively. He worked in the UVA Microwaves and Semiconductor Devices Laboratory between 1996-2001 as a research assistant. Later he worked in UVA Microwaves Lab as a Research Associate for one year. He joined Girne American University, Girne, Cyprus, in September 2002. He was a member of electrical and electronics engineering department as an Assistant Professor until 2009. Since 2009, he has been promoted to Associate Professor status and he has also been the vice-rector of the Girne American University responsible from academic affairs. His research interests include Microwave Active Circuits, Microwave Measurement Techniques, Numerical Methods, and Metaheuristic Algorithms and their applications to design problems. He has authored and coauthored technical publications in the areas of submillimeter wavelength measurements and particle swarm optimization technique. He is a member of IEEE and Eta Kappa Nu. He is also the recipient of Louis T. Rader Chairpersons award in 2001.